\documentclass[12pt]{article}

\usepackage{newtxtext,newtxmath}
\usepackage{graphicx}
\usepackage[letterpaper,margin=1in]{geometry}

\renewenvironment{abstract}
	{\quotation}
	{\endquotation}

\date{}

\makeatletter
\renewcommand{\fnum@figure}{\textbf{Figure \thefigure}}
\renewcommand{\fnum@table}{\textbf{Table \thetable}}
\makeatother

\usepackage{scicite}

\usepackage{url}

\def\scititle{
    Robots that learn to evaluate models of collective behavior
}

\title{\bfseries \boldmath \scititle}

\author{
	\parbox[t]{0.92\textwidth}{\centering
		Mathis~Hocke$^{1,\ast}$,
		Andreas~Gerken$^{1}$,
		David~Bierbach$^{2,3,4}$,
		Jens~Krause$^{2,3,4}$,
		Tim~Landgraf$^{1,\ast}$}\and
	\small$^{1}$Department of computer science, Freie Universität Berlin, 14195 Berlin, Germany.\and
	\small$^{2}$SCIoI Excellence Cluster, Technische Universität Berlin, 10587, Berlin, Germany.\and
	\small$^{3}$Faculty of Life Sciences, Humboldt-Universität zu Berlin, 10117, Berlin, Germany.\and
	\small\parbox[t]{0.92\textwidth}{\centering$^{4}$Department of Fish Biology, Fisheries, and Aquaculture, Leibniz Institute of Freshwater Ecology and Inland Fisheries, 12587, Berlin, Germany.}\and
	\small\parbox[t]{0.92\textwidth}{\centering$^\ast$Corresponding authors. Email: mathis.hocke@fu-berlin.de, tim.landgraf@fu-berlin.de}\and
}

\begin{document} 

\maketitle

\begin{abstract} \bfseries \boldmath

Collective animal behavior depends on how individuals respond to one another and to their surroundings, and computational models are central to explaining these interactions.
Biomimetic robots make these interactions experimentally accessible by acting as controllable social partners whose behavior can be specified and varied in real time.
However, animal-behavior models are commonly evaluated by comparing simulated and recorded trajectories offline, leaving unclear whether they predict behavior during closed-loop interaction.
Here we show that reinforcement learning with a biomimetic robotic fish can be used as an embodied test of candidate models of fish behavior.
For each of four fish models---a constant-following baseline, two rule-based models and a convolutional neural network model---we trained a robotic policy in simulation to lead a simulated fish between goal locations, and then deployed the policy on a real robot interacting with live fish.
We quantified model fidelity as the sim-to-real gap between simulated and real interaction distributions, including goal reaching, inter-individual distance, alignment, wall interactions and locomotion.
Across task performance and additional behavioral measures, these gaps differed systematically among candidate models and exposed discrepancies that offline trajectory comparisons can miss.
The closest-matching candidate still left residual gaps, especially in individual variability, showing that the benchmark can diagnose model limitations as well as rank candidate models.
Learned robotic policies therefore act as embodied probes that extract and test the predictions of behavioral models in live interaction.
This approach suggests a general route for testing computational models not only by how well they reproduce past data, but by whether they support successful interaction with the real systems they seek to explain.

\end{abstract}

\noindent
Biomimetic robots have become powerful instruments for probing social interactions in animal groups. By replacing one or more individuals with controllable robotic replicas, researchers can directly test how animals perceive, respond to, and coordinate with conspecifics \cite{krause_interactive_2011, landgraf_interactive_2013}. Such experiments are particularly valuable for collective behaviors such as fish shoaling, collective decision making and abstract communication, where simple local interactions among individuals give rise to complex group-level dynamics \cite{vicsek_novel_1995, couzin_collective_2002, halloySocialIntegrationRobots2007, landgrafDancingHoneyBee2018c}.

Understanding these emergent phenomena requires more than reproducing observed trajectory statistics; it requires the full scientific cycle of observation, model building, hypothesis generation, experimental intervention, and renewed observation. Mathematical models—ranging from zone-based frameworks incorporating repulsion, alignment, and attraction \cite{aoki_simulation_1982, reynolds_flocks_1987, huth_simulation_1992} to vision-based and machine-learning approaches \cite{eyjolfsdottir_learning_2016, heras_deep_2019, costa_automated_2020}—provide candidate interaction rules, but fitting such models to recorded trajectories remains largely correlational. A model may match observed patterns while still failing to capture the causal response dynamics that unfold during social interaction.

Biomimetic robots are valuable in this context because they enable time-resolved, closed-loop perturbations of social behavior. In closed loop, the animal's behavior immediately alters the robot's next action, allowing hypothesized interaction rules to be tested under intervention rather than passive playback. Existing robot studies have used this capability mainly to establish social acceptance by animal groups \cite{faria_novel_2010, landgraf_robofish_2016} or to test isolated behavioral cues under open-loop or narrowly targeted closed-loop conditions \cite{polverino_fish_2013, bierbach_using_2018, maxeiner_social_2023}. What is still missing is a framework that uses such embodied interactions to evaluate full computational models rather than isolated cues.

Relatedly, biohybrid robot–fish studies have used closed-loop experiments to assess the fidelity of transferring learned interaction models from simulation to embodied robots, quantifying a “biomimicry gap” by comparing behavioral observables across fish–fish pairs, fish–robot pairs, and simulations \cite{papaspyrosQuantifyingbiomimicrygap2024}. Our objective differs: rather than asking whether a particular interaction model can be executed on a robot without introducing artifacts from embodiment and signaling, we use the robot as an embodied hypothesis test. For each candidate behavioral model, we train a robot control policy that should succeed if that model captures the relevant interaction dynamics; comparing simulated and live-fish responses under the same policy then provides the empirical test.

A related challenge arises for purely data-driven models: deep-learning approaches can reproduce observed trajectory statistics with remarkable fidelity \cite{papaspyrosQuantifyingbiomimicrygap2024}, but because they are fit to data rather than derived from mechanistic principles, they cannot be falsified in the classical Popperian sense---a sufficiently flexible model may match any dataset without embodying a testable causal hypothesis. What is missing, then, is a methodology that mirrors the two core steps of the scientific method applied to behavioural models: first, extract \emph{testable predictions} from a candidate model by training an agent whose strategy should succeed if the model is correct; and second, \emph{test those predictions empirically} by deploying that agent in a real animal--robot interaction experiment.

The missing step is an experimental loop in which a candidate model is first converted into an actionable prediction and then tested under the same closed-loop conditions that define social interaction. Biomimetic robots uniquely enable active hypothesis testing in natural systems, while models provide testable abstractions of interaction rules.

\begin{figure}
	\centering
	\includegraphics[width=0.8\textwidth]{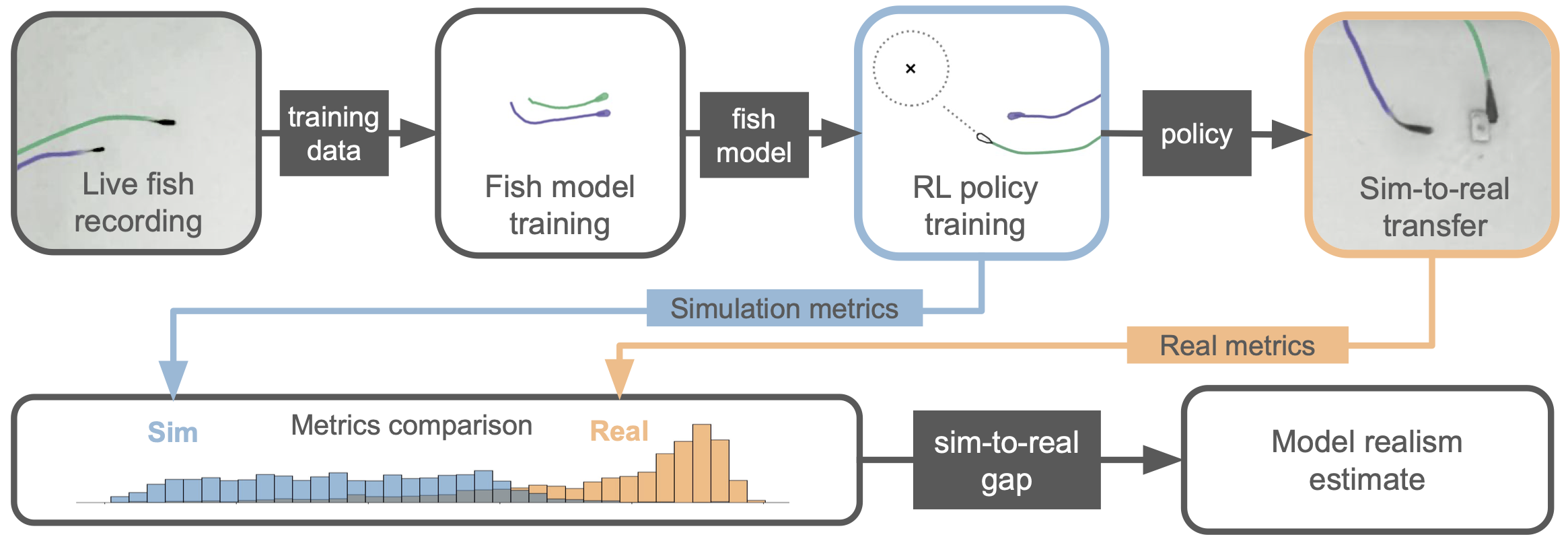}
	\caption{\textbf{Workflow of our sim-to-real model evaluation framework.}
     Live fish trajectories are recorded and used to train behavioral models. Each model is then used to train an RL control policy in simulation. The learned policy is transferred to real fish–robot interaction experiments, behavioral metrics are compared between simulation and reality, and the resulting sim-to-real discrepancy produces a model-realism estimate.
    }
	\label{fig:visual_abstract}
\end{figure}

Here, we bridge this gap by integrating biomimetic robotic fish with reinforcement learning (RL) to evaluate and validate models of individual behavior in fish shoals. In our framework, a robotic control policy is trained in simulation in an environment with virtual fish governed by one of several behavioral models—ranging from rule-based to data-driven formulations. The learned policy is then transferred to the real robotic platform and tested with live guppies (\textit{Poecilia reticulata}). In this sense, each learned policy operationalizes an embodied hypothesis about the interaction strategy implied by the underlying behavioral model. The difference between simulated and real-world performance, quantified as the simulation-to-reality (sim-to-real) gap, provides a direct, quantitative test of that hypothesis and a quantitative measure of model realism (Fig.~\ref{fig:visual_abstract}).

Sim-to-real transfer is well established in robotics and autonomous driving \cite{tobin_domain_2017, peng_sim--real_2018}.
It has also been used to compare simulator models outside animal behavior: one cloth-manipulation benchmark evaluated four deformable-object simulators by replaying predefined dynamic and quasi-static manipulation trajectories and comparing simulated cloth meshes with real-world point clouds \cite{blanco-mulero_benchmarking_2024}.
This shows that simulator fidelity can be assessed by sim-to-real gap under standardized open-loop rollouts.
Our framework goes beyond this setting by training a dedicated closed-loop policy for each candidate model and transferring each policy to a real animal--robot interaction, where every action depends on the live animal's response.
Although robot–animal interactions have long been used to test behavioral hypotheses, and biohybrid systems have been used to quantify transfer-related discrepancies when deploying interaction models on robots, sim-to-real transfer has not been formalized as a model-agnostic framework for ranking behavioral models by comparing live-animal responses to simulated-model responses under dedicated closed-loop probing policies.
We use sim-to-real transfer to turn robotic interaction into a quantitative test of model realism. This framework provides a generalizable method for comparing and ranking behavioral models based on their ability to reproduce real animal–robot interactions, thereby advancing both modeling and experimental research on collective animal behavior by demonstrating that closed-loop model evaluation can be made quantitative. We instantiate this framework with a leadership-to-goal task, because collective movement toward destinations such as food sources or home routes is a classical problem in animal-behavior research \cite{krauseLeadershipfishshoals2000, couzinEffectiveLeadershipDecisionmaking2005, strandburg-peshkinInferringInfluenceLeadership2018a} and provides a simple objective with a clear quantitative readout in dyadic fish--robot interactions.

\subsection*{Robot–Fish Interactions Expose Model Gaps}

To instantiate the benchmark, we selected four candidate fish models spanning a simple baseline, mechanistic rule-based formulations, and a data-driven model. The baseline model $M_{Follow}$ served as a minimal reference, $M_{Zone}$\cite{couzin_collective_2002} and $M_{Force}$\cite{klamser_impact_2021} represented two established rule-based approaches with distinct interaction assumptions, and $M_{ConvNet}$ provided a neural model trained on guppy trajectories. The labels $M_{Zone}$ and $M_{Force}$ are compact descriptive shorthand: $M_{Zone}$ denotes the model whose interaction rules are defined by three distance-dependent behavioral zones (repulsion, orientation, and attraction), whereas $M_{Force}$ denotes the model whose force-based equations of motion represent self-propulsion, velocity alignment, and distance regulation as social force terms acting on individual motion. Together, these models form a deliberately heterogeneous test bed for the evaluation framework rather than an exhaustive comparison of model families.

For each model, we then trained a dedicated RL policy--$\pi_{Follow}$, $\pi_{Zone}$, $\pi_{Force}$, and $\pi_{ConvNet}$--to control the robot in simulation. The task was to lead the fish through a sequence of goal locations, and we used the number of goals reached by the fish ($N_{goals}$) as the primary performance measure. All policies successfully learned the task in simulation and developed distinct model-specific strategies.

See Materials and Methods in the supplementary materials \cite{methods} for more details on the models.

\subsubsection*{The Benchmark Separates Models}

\begin{figure}
	\centering
	\includegraphics[width=0.6\textwidth]{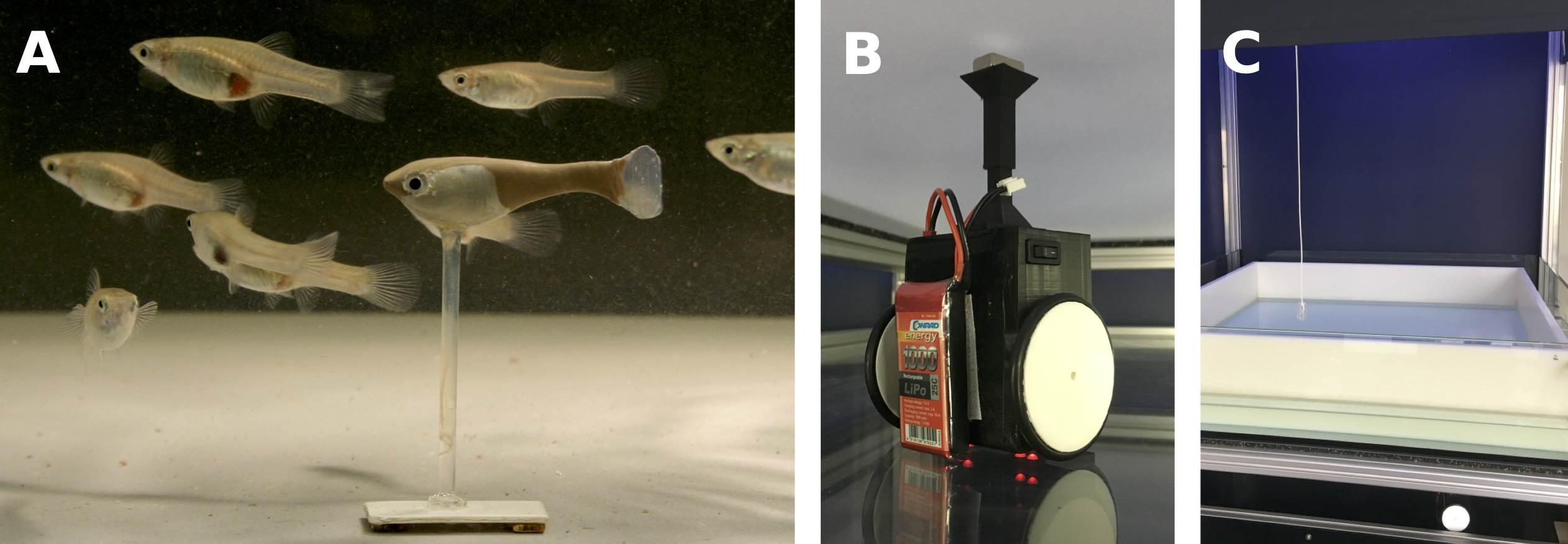}
	\caption{\textbf{Experimental setup for guiding live guppies with a magnetically controlled replica.}
    (\textbf{A}) Live guppies interacting with the magnetically controlled guppy replica in the tank.
    (\textbf{B}) Two-wheeled robot positioned below the tank that controls the replica's movement.
    (\textbf{C}) 1$\times$1~m experimental tank with the robot underneath.
    }
	\label{fig:setup}
\end{figure}

\begin{figure}
	\centering
	\includegraphics[width=0.4\textwidth]{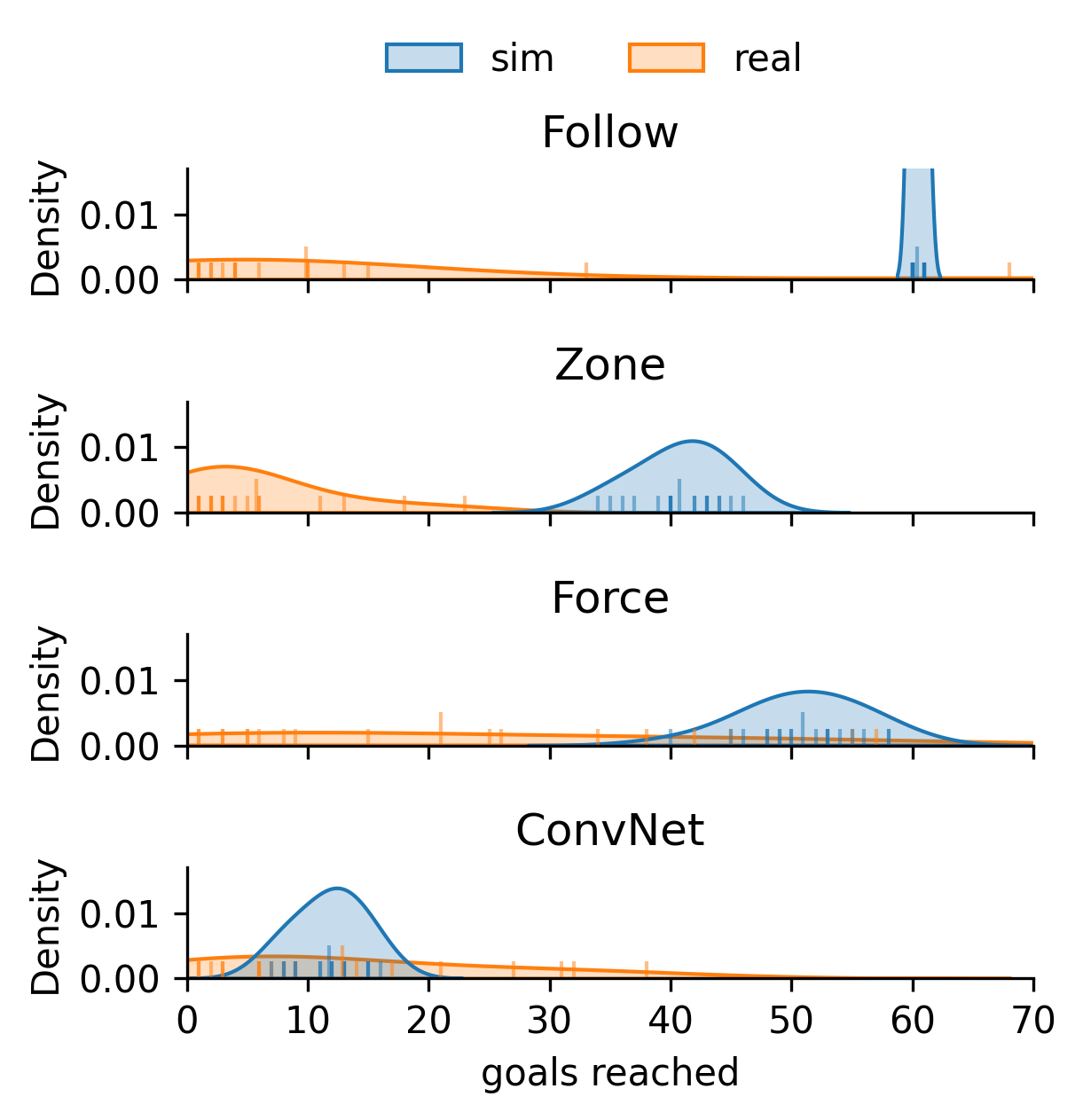}
	\caption{\textbf{Distributions of the number of goals reached.}
        Goals reached in simulation (blue) and in the real environment with a live guppy (orange). Models $M_{Follow}$, $M_{Zone}$ and $M_{Force}$ allow the respective policies to learn strategies that perform much better in simulation than they do with real fish. The policy trained with $M_{ConvNet}$ performs more similarly in simulation and reality. The distributions in the real setup are consistently wider than in simulation. Each distribution is based on $n=18$ trials per policy and environment. Total goals across all trials (simulation/real): $\pi_{Follow}$ 1087/177, $\pi_{Zone}$ 733/104, $\pi_{Force}$ 917/378, $\pi_{ConvNet}$ 212/232.
    }
	\label{fig:goals}
\end{figure}

We deployed all policies on the real RoboFish platform, in which a 3D-printed guppy replica inside the tank is magnetically coupled to a two-wheeled robot moving beneath the tank (Fig.~\ref{fig:setup}).
We conducted 18 trials per policy (15~min each) with live female guppies.
In comparison to simulation, three policies—$\pi_{Follow}$, $\pi_{Zone}$, and $\pi_{Force}$—achieved fewer goals in the real environment, whereas $\pi_{ConvNet}$ showed a slight increase. The distributions of $N_{goals}$ were consistently broader in real experiments than in simulation, i.e.\ animals showed a high degree of inter-individual variation (Fig.~\ref{fig:goals}).

We quantified the sim-to-real gap as the Wasserstein distance between simulated and real trial-level $N_{goals}$ distributions (95\% percentile-bootstrap CI).
The gap was highest for the policy trained with baseline model $M_{Follow}$ (51.3; 95\% CI [44.5, 56.6]), followed by the policies trained with rule-based models $M_{Zone}$ (34.9; 95\% CI [31.3, 37.9]) and $M_{Force}$ (29.9; 95\% CI [21.2, 38.6]).
The policy trained with $M_{ConvNet}$ exhibited the smallest gap (7.8; 95\% CI [5.3, 11.0]).
Permutation tests confirmed significant differences between simulation and reality for all policies (all $p < 0.01$), with large effect sizes for $\pi_{Follow}$ (Cliff's $\delta = 0.89$), $\pi_{Zone}$ ($\delta = 1.0$), and $\pi_{Force}$ ($\delta = 0.8$), and a smaller effect for $\pi_{ConvNet}$ ($\delta = 0.21$).
Under this benchmark, that policy therefore produced the closest match to real fish–robot interactions, but the more general result is that the evaluation separates the candidate models in a consistent, quantitative way while still exposing a substantial residual gap even for the best match.

\paragraph*{The sim-to-real gap is model-induced, not noise.}
To confirm that these differences are genuinely induced by the fish models rather than by optimization noise, we performed simulation-to-simulation transfer experiments (``sim-to-sim''), in which each policy interacted with each fish model. Policies consistently achieved the highest performance on the model they were trained with, and performance dropped markedly when transferred to another model (Fig.~\ref{fig:simtosim}). We estimated the sim-to-sim gap as the median Wasserstein distance of $N_{goals}$ (with 95\% bootstrap CI) for each policy–model pair.
When the same model was used for rollout as during training, the gap remained very low ($\leq 1.22$); when a different model was used, the gap was substantially larger (7.90–47.54).
Kruskal–Wallis omnibus tests revealed strong multi-group separation across rollout models for each trained policy ($\pi_{Follow}$: $H=153.49$, $p=4.65 \times 10^{-33}$; $\pi_{Zone}$: $H=180.73$, $p=6.14 \times 10^{-39}$; $\pi_{Force}$: $H=184.93$, $p=7.61 \times 10^{-40}$; $\pi_{ConvNet}$: $H=163.89$, $p=2.66 \times 10^{-35}$).
Planned pairwise permutation tests between each policy's training model and all alternatives were significant after Holm correction (all $p < 0.001$), demonstrating that each model induces a distinct interaction regime and that policies do not converge on generalizable solutions.

We further tested whether random-seed variation during training could explain the observed differences.
Five additional policies per model were trained with identical hyperparameters but independent seeds (six policies per model total).
Within-model policy variance was far smaller than between-model variance (variance ratio $R = \sigma^2_{\mathrm{between}}/\sigma^2_{\mathrm{within}} = 133.79$; 95\% bootstrap CI $[93.87, 171.18]$; permutation test $p<0.0002$), confirming that the benchmark rankings are dominated by model-induced interaction dynamics rather than training stochasticity.

\subsubsection*{Per-Metric Diagnostics Reveal Model Artifacts}

Having established that the benchmark reliably separates models, we next asked \textit{how} these models differ. We extended the goals-reached analysis to nine per-time-step behavioral metrics—inter-individual distance (IID), IID change, alignment, fish speed, fish turn, fish–goal distance, fish-faces-robot, wall distance, and wall alignment—computing trial-level Wasserstein distances and summarizing policy rankings per metric in an annotated heatmap (Fig.~\ref{fig:gap}).
Metric definitions and trial-level gap computations are detailed in Materials and Methods in the supplementary materials \cite{methods}.
Averaging ranks across all per-time-step metrics yielded mean ranks of $\pi_{Follow}$: 3.56, $\pi_{Zone}$: 3.11, $\pi_{Force}$: 2.11, and $\pi_{ConvNet}$: 1.22, consistent with the ranking induced by the goals-reached metric alone.

Occasional rank inversions were informative rather than contradictory, because they revealed specific model artifacts. The IID-change metric penalized model $M_{Zone}$ because its sharp zone boundaries produce abrupt, unrealistic fluctuations in inter-individual distance. Conversely, the characteristic behavior of $\pi_{Follow}$—moving quickly to the goal and waiting—produced IID-change distributions that coincidentally resembled those of real fish. Model $M_{Zone}$ also exhibited unrealistically strong alignment: because the robot frequently positioned itself in the model's zone of orientation (ZOO), sustained alignment responses exceeded levels observed in live guppies. The relatively high turning noise in model $M_{Force}$ may have contributed to its small gap on several metrics, but it resulted in a large sim-to-real gap for the fish-turn metric specifically.

\paragraph*{Approach behavior influences the IID differences.}
The underlying mechanism driving many of these metric differences is the distinct approach behavior encoded by each policy. To isolate attraction dynamics from the confounds of a moving interaction partner, we performed dedicated simulation trials in which each policy interacted with a stationary fish. We measured (i) the minimum robot–fish distance and (ii) the proportion of time the robot spent closer to the fish than to the goal.
Policies differed markedly (Fig.~\ref{fig:approach}): $\pi_{Follow}$ did not approach the fish at all; $\pi_{Zone}$ and $\pi_{Force}$ frequently approached within 4~cm and 0.35~cm respectively, spending 22.3\% and 56.1\% of episode time closer to the fish than to the goal; $\pi_{ConvNet}$ exhibited an intermediate strategy, maintaining a larger average distance (29~cm) and spending only 4.6\% of the time closer to the fish.

\begin{figure}
	\centering
	\includegraphics[width=0.9\textwidth]{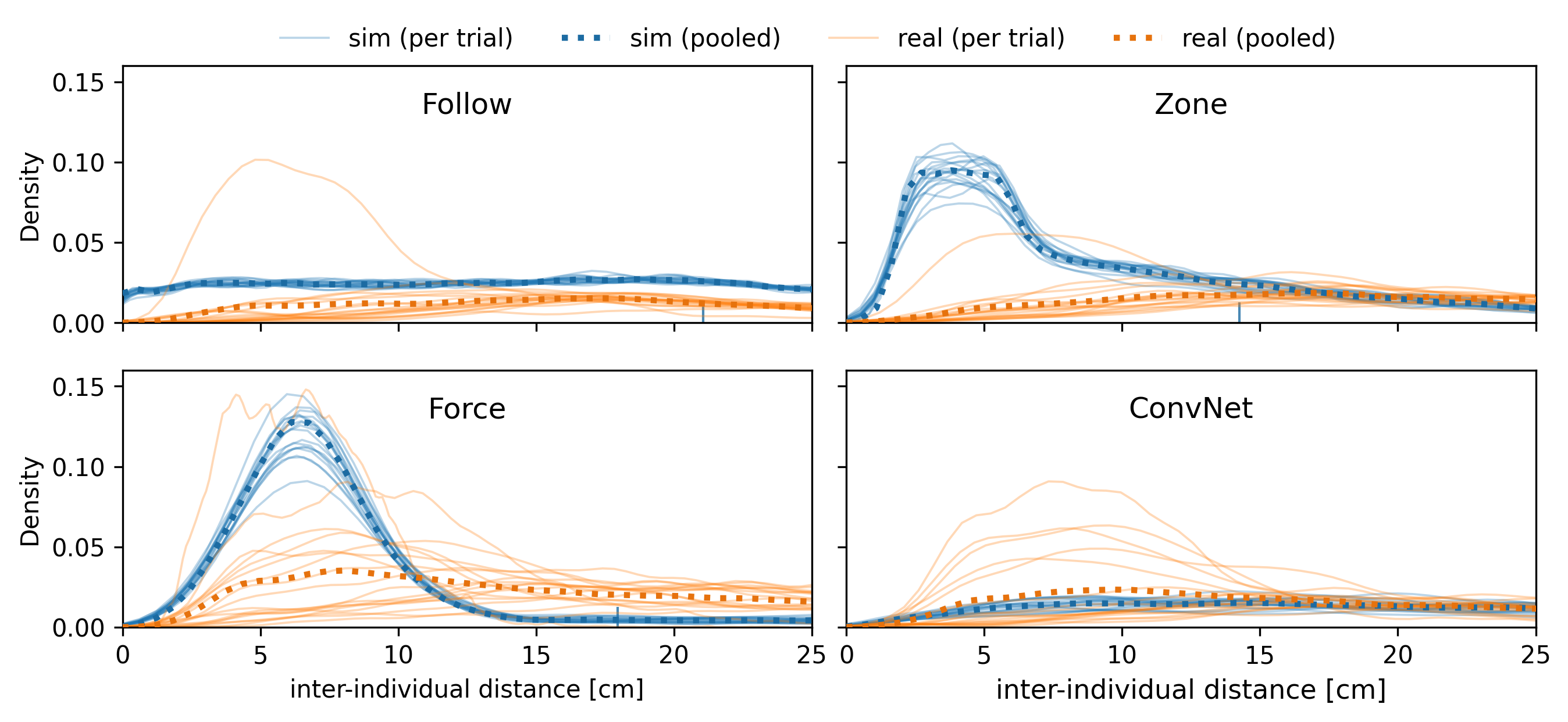}
    \caption{\textbf{Inter-individual distance (IID) in simulation (blue) and in the real environment with a live guppy (orange).}
    Thin solid lines show per-trial kernel density estimates over per-time-step IID samples ($n=18$ trials per policy and environment; 15 min at 25 Hz), highlighting inter-trial variability. Dotted lines indicate the pooled IID distributions across all trials.
    }
	\label{fig:iid}
\end{figure}

These approach regimes help explain the inter-individual distance distributions observed in live experiments (Fig.~\ref{fig:iid}). Three models ($M_{Follow}$, $M_{Zone}$, $M_{Force}$) maintained unrealistically tight proximity to the robot in simulation, whereas $M_{ConvNet}$ generated a broader IID distribution that more closely matched real fish behavior. Examining trial-level IID distributions reveals substantial inter-individual variability in the real experiments: trials in which fish followed the robot closely exhibited pronounced peaks at low IID values, while other trials showed substantially broader distance distributions. Policies trained with models $M_{Zone}$ and $M_{Force}$ produced strong low-IID peaks resembling those of the closely following outlier fish, whereas the policy trained with $M_{ConvNet}$ matched the IID distributions observed in the majority of trials but did not reproduce the tightly peaked distributions characteristic of strong-following outliers. Similarly, the policy trained with $M_{ConvNet}$ most closely reproduced the guppies' tendency to swim near walls and follow tank boundaries.

\subsubsection*{Models Share Common Weaknesses}

While the analyses above highlight how the benchmark \textit{separates} models, a complementary finding is that certain behavioral features expose shared limitations across all four model classes.
Across all models, simulations also overestimated the proportion of time the fish oriented toward the robot.

\paragraph*{Speed and locomotion style.}
Speed distributions reflected inherent constraints of the models: $M_{Follow}$ and $M_{Zone}$ use constant-speed formulations and therefore cannot reproduce the burst-and-coast variability of live guppies. None of the models captured the extended motionless periods characteristic of real fish.

\paragraph*{Time-to-goal inflation.}
Average time to goal, defined as the time from a goal becoming active to that goal being reached and averaged over reached goals within a trial, differed substantially between simulation and reality. In simulation, goals were reached after
14.8~±~1.1~s ($\pi_{Follow}$), 21.8~±~12.2~s ($\pi_{Zone}$), 17.4~±~13.0~s ($\pi_{Force}$), and 72.0~±~58.3~s ($\pi_{ConvNet}$).
In real experiments, times to goal increased to
52.1~±~82.7~s, 101.2~±~149.2~s, 36.3~±~61.9~s, and 60.9~±~95.4~s, respectively.
These results indicate that the models fail to capture not only the variability in the number of goals reached per trial, but also the variability in time to goal.

\begin{figure}
	\centering
	\includegraphics[width=0.6\textwidth]{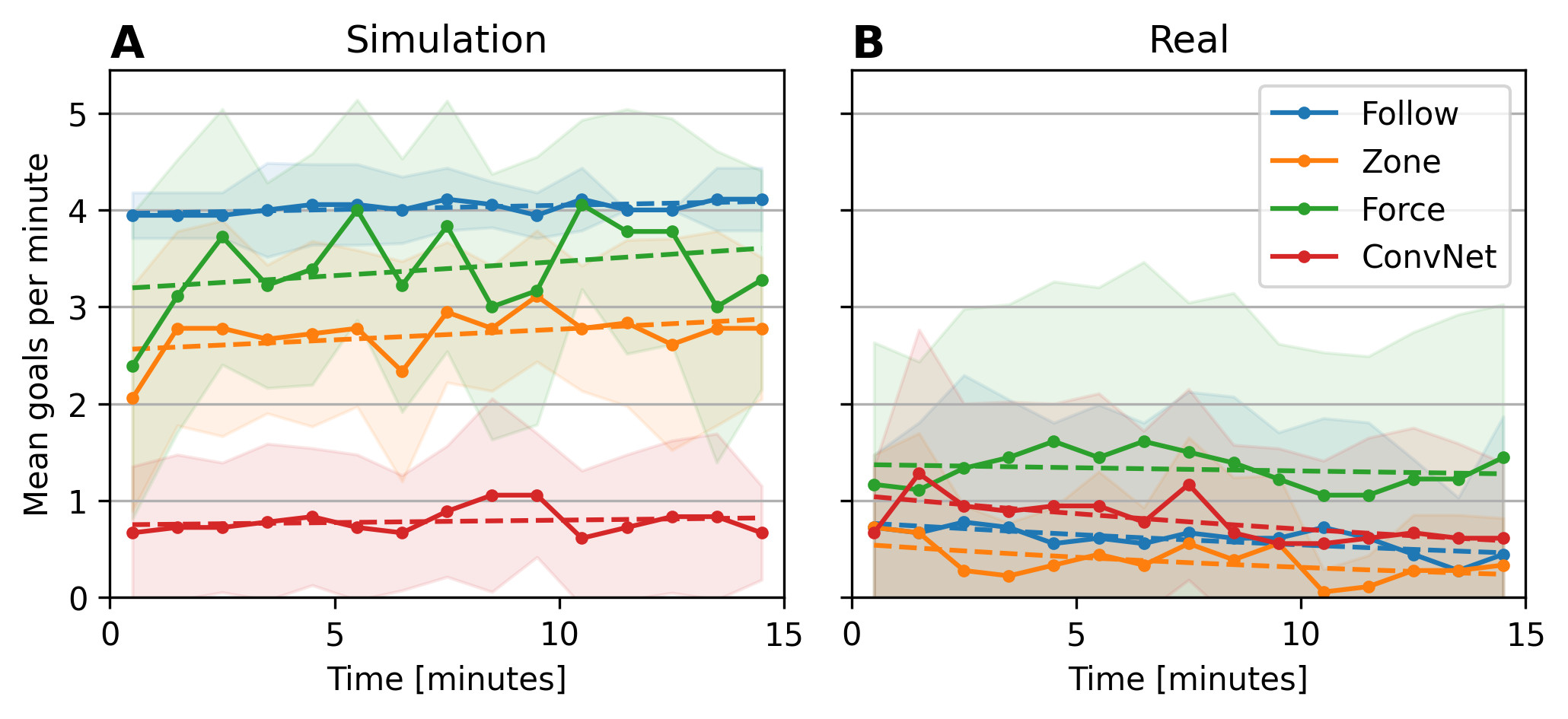}
    \caption{\textbf{Mean goals reached per minute over time per policy in simulation and with live guppies.}
    (\textbf{A}) Simulation results. (\textbf{B}) Results with live guppies. Lines show the mean across $n=18$ trials per policy; shaded areas show one standard deviation. Each trial starts with two minutes of milling, followed by 15 minutes of closed-loop control shown in the plot. Dashed lines indicate fitted negative-binomial regression trends for per-minute goal counts. See Appendix Table~\ref{tab:temporal} for regression results.
    }
	\label{fig:goals_reached_over_time}
\end{figure}

\paragraph*{Temporal decline in engagement.}
Each trial lasted 15~min, as guppies typically reduce interaction toward the end of longer sessions. In real experiments, goals reached per minute showed a small but statistically significant decline over the trial (rate ratio per minute $=0.974$; $p=0.005$; Table~\ref{tab:temporal}). The effect is consistent with expected fatigue but remained shallow in practical terms, and goal reaching persisted throughout the full 15-minute window, so we retained full-length trials for all analyses. In simulation, goal rate showed a slight increase over time (rate ratio per minute $=1.006$; $p=0.004$), consistent with an early-trial transient rather than any within-trial state change. Variance between policies was lower in simulation but higher across policies in reality, underscoring the greater diversity of behavioral outcomes in live interactions (Fig.~\ref{fig:goals_reached_over_time}).

\paragraph*{Inter-individual variability.}
Across all metrics, simulated distributions were consistently narrower than their real counterparts. Capturing this variability will require models that represent not just an average fish but a distribution of individuals with different traits or internal states—a challenge none of the tested model classes currently addresses.

\subsection*{Closed-Loop Probes Test Model Realism}

Our results demonstrate that sim-to-real transfer of robotic interaction policies provides a rigorous and quantitative lens for evaluating behavioral models in collective animal behavior. By training policies in simulation with different candidate models and subsequently deploying them on a physical biomimetic robot interacting with live guppies, we were able to directly measure how well each model captures real fish–robot interaction dynamics. Across four tested models, the policy trained with $M_{ConvNet}$ consistently produced the smallest sim-to-real gap across most behavioral metrics, indicating that the data-driven model most accurately reproduced natural fish behavior in our pairwise leadership task. More importantly, this ranking demonstrates that the proposed framework can distinguish candidate models in a principled, closed-loop manner.

A central strength of the framework is that each policy serves not only as an implicit probe for the interaction rules encoded by its underlying fish model, but also as a testable hypothesis about how a robot should act if that model were correct. In simulation, policies converged on distinct strategies—ranging from never approaching the fish ($\pi_{Follow}$), to approaching but maintaining moderate distance ($\pi_{ConvNet}$), to repeatedly approaching very closely ($\pi_{Zone}, \pi_{Force}$)—and these strategies transferred to the real robot. Sim-to-sim transfer experiments further revealed that these strategies were model-specific: replacing a model's dedicated policy with one trained on a different model consistently degraded performance. Independent-seed retraining showed that benchmark rankings were dominated by between-model differences rather than by training stochasticity. Together, these findings highlight that policies encode a behavioral “fingerprint” of the model they were trained on, providing a unique tool for assessing model similarity and behavioral dynamics that are difficult to extract analytically.

This emphasis on model-specific policies distinguishes our benchmark from traditional simulator comparisons based on fixed rollouts. In open-loop settings, all models are evaluated under the same predefined perturbations. Our framework goes beyond this setting by training a dedicated closed-loop policy for each candidate model and transferring each policy to the real system. Each policy is therefore shaped by the model it was trained with, exploiting its characteristic response dynamics; at deployment, the closed-loop controller conditions every action on the real fish's behavior in real time rather than replaying a fixed trajectory.

This principle is not specific to animal behavior. Whenever multiple candidate models can be coupled to an embodied system and probed through robotic interaction, one can train a policy against each model and evaluate the models by the resulting sim-to-real gap. This policy-based evaluation also avoids a form of leakage that can arise in fixed-rollout benchmarks: if a model is tuned to minimize its gap to the same real-world trajectories used for evaluation, the resulting score may reflect overfitting to that benchmark rather than generalizable model realism. In our workflow, the candidate model is fixed before policy training, and the evaluation is the live response to a learned closed-loop probe rather than a replay of the trajectories used to fit the model. Because RL policies optimize against the simulator they are trained in, they can discover and exploit subtle model artifacts that fixed trajectories may not expose.

The baseline model $M_{Follow}$ offers a useful reference point for interpreting these results. Because the model moves at a constant speed and always orients toward the robot, the optimal solution for the policy is to move from goal to goal and wait—requiring no attention to the fish at all. This trivial strategy performs surprisingly well even for live guppies, underscoring the importance of comparing against simple baselines. The fact that all policies outperformed $\pi_{Follow}$ in sim-to-real transfer suggests that the remaining models encode interaction dynamics that the policy can exploit in a meaningful way.

An important distinction is that success with live fish is not, by itself, a measure of model realism. A policy may reach many goals with guppies because it discovers a generally effective robotic tactic, while the model that produced that tactic still gives an unrealistic response. $M_{Follow}$ makes this point explicitly: it was designed around attraction, a cue already known to be useful for the robot, and therefore leaves $\pi_{Follow}$ no choice but to rely on attraction. The resulting policy reaches a nontrivial number of goals with live fish, but the model remains a poor description of social behavior; for example, a swarm of $M_{Follow}$ agents would simply collapse to a point. The same distinction explains why $\pi_{Force}$, although it reached the most goals with live fish, is not ranked as the most realistic model. Its policy also reached many goals in simulation, so the simulated response was still much stronger than the live-fish response. By contrast, $\pi_{ConvNet}$ reached fewer goals with live fish, but its simulated performance was much closer to the live-fish mean. Thus, the relevant criterion is not which learned strategy maximizes real-world task performance, but whether the behavior elicited by that strategy is similar in the model and in live fish. In this sense, $M_{Force}$ is informative rather than simply wrong: it supports a strategy that also works with guppies, but its response is exaggerated, much like the attraction bias in $M_{Follow}$.

The rule-based models tested here, $M_{Zone}$ and $M_{Force}$, represent only a single parameterization of each model, and these parameter sets were not tuned for guppy behavior or for pairwise robot-fish interactions.  Their limited performance should therefore not be taken as a critique of their modeling principles.  Instead, our approach offers a systematic means to evaluate future parameterizations, model variants, or entirely new rule-based formulations. Moreover, these models were originally designed for group contexts, whereas our experiments involved only two agents, the robot and a single fish; multi-fish scenarios may reveal strengths of rule-based models that remain hidden in dyadic settings. As larger-group robotic experiments become feasible, our framework can be extended to evaluate models at the group level, including those trained on group trajectories.

Noise in model dynamics played a notable role in policy training. While increasing the turning noise of the rule-based models made the policy's task less predictable and slowed learning, it also expanded the set of reachable states, forcing the policy to recover from situations that never arise in perfectly deterministic models - comparable to the effects of domain randomization. This improved robustness and broader state coverage are particularly important for sim-to-real transfer, where unpredictable deviations are inherent to live animal behavior. Noise thus served as a substitute for missing behavioral richness, partially compensating for the models' limited complexity.

The robot's action space—restricted to selecting a desired turning direction at 1 Hz—was intentionally kept simple to enable fast training and stable control. These high-level commands were executed through a PID-based motion controller that produced guppy-inspired burst movements. While this design simplifies training and reduces hardware variability, it also constrains the set of interaction behaviours the policy can express. Future work could explore lower-level control or end-to-end learning of velocity profiles, enabling richer interactions and potentially uncovering model-specific behavioral features that remain inaccessible with high-level actions.

An additional limitation arises from the robot's actuation constraints. While the robot can match and exceed the guppies' average swimming speed, it cannot reproduce their brief high-acceleration bursts, partly due to mechanical limits of the magnetic coupling between the robot and the submerged replica. As a result, learned policies are biased toward attraction-based strategies that lead the fish toward the goal, whereas shepherding behaviors are likely more difficult to realize. Although these constraints were faithfully mirrored in simulation, relaxing them in silico could reveal alternative interaction strategies that remain unattainable in the current physical setup, highlighting how hardware limitations shape the space of learnable robot behaviors.

Our evaluation metric, the Wasserstein distance between simulated and real distributions of behavioral measures, generally revealed consistent model rankings across metrics, with occasional metric-specific deviations. The number of goals reached emerged as a reliable proxy for the overall fidelity of fish-robot interaction, largely mirroring rankings obtained from a broader set of behavioral distributions. Occasional rank inversions for specific metrics were informative rather than contradictory: they illuminated particular weaknesses in model dynamics—for example, an unrealistic alignment in model $M_{Zone}$ or a coincidental match in IID-change for model $M_{Follow}$ due to policy waiting behavior. Such divergences illustrate how the framework not only produces an overall ranking but also diagnoses model-specific behavioral artifacts.

Across most metrics, $M_{ConvNet}$ achieved the smallest sim-to-real gap. Its superior performance is notable given the covariate shift between training and deployment. $M_{ConvNet}$ was trained on homogeneous pairs of female guppies, whereas during testing it interacted with a robot executing policy-driven movements that differ substantially from natural guppy kinematics. $M_{ConvNet}$ therefore encountered interaction regimes far outside its training distribution. Its robust performance suggests that key features of guppy interaction—such as distance-dependent responsiveness and wall-following tendencies—were captured in a transferable manner.
We stress, however, that this is not a claim that $M_{ConvNet}$ is the definitive model of guppy behavior. Rather, the importance of this result is that the evaluation can identify a comparatively strong candidate and expose where alternative models fail under the same embodied probe.
At the same time, interpretability remains a distinct scientific criterion. A model that matches behavior without revealing which interactions or cues drive that match offers limited mechanistic insight; however, the policies learned against a model provide an additional explanatory layer. Their closed-loop strategies expose which parts of the model can be used to elicit behavioral change, turning model assumptions into observable interaction hypotheses that can be inspected experimentally.

In this sense, the framework operationalizes Popperian falsification for behavioral models. Each learned policy encodes a testable prediction—the interaction strategy implied by the underlying model—and the sim-to-real gap delivers the empirical verdict. A model whose policy fails to reproduce real interaction dynamics is falsified under the tested conditions. Crucially, even the best-performing data-driven model exhibited a significant residual gap, demonstrating that embodied closed-loop testing can reveal limitations that are difficult to detect from offline trajectory fitting alone. This provides an additional form of empirical validation, especially for data-driven approaches that are otherwise assessed mainly against held-out trajectory data.

A major discrepancy between simulations and real experiments was the variability in fish behavior. All models produced narrow distributions of e.g. goals reached across episodes, whereas live guppies exhibited substantially greater variability between individuals. 
Likewise, while simulated trials yielded highly consistent distributions of metrics such as IID, the real experiments showed substantial trial-to-trial variability, with individual outliers present for every tested policy.
Capturing this variability will require models that represent not just an average fish but a distribution of individuals with different traits or internal states. Extending behavioral models to reflect inter-individual variability is therefore a key step toward reducing the sim-to-real gap further and toward developing artificial agents capable of interacting fluently with diverse animals.

The benchmark also exposed shared deficiencies in locomotion and temporal realism. The constant-speed formulations could not reproduce the burst-and-coast structure of guppy movement, and none of the models captured the extended pauses seen in live fish. Likewise, simulated interactions generally produced shorter and less variable times to goal than the real trials, and they did not reproduce the shallow decline in engagement observed over time in the live experiments. These mismatches suggest that future behavioral models must account not only for interaction rules, but also for internal state, movement style, and time-varying responsiveness.

Finally, our framework opens new opportunities for designing task objectives that probe specific behavioral hypotheses. While the present study focused on a simple leadership-to-goal task suited for dyadic interactions, future objectives could influence alignment, separation, group cohesion and decision making, or the propagation of movement cues in multi-fish settings. An especially promising direction is an out-of-distribution (OOD) objective, in which a policy is rewarded for driving the interaction into regions where the fish model's inputs are least supported by its training distribution, so that sim-to-real transfer preferentially collects novel data for iterative model and policy retraining, gradually filling gaps in the behavioral dataset. Such an objective would already instantiate a form of automated science: the policy implicitly proposes hypotheses about where the model is underspecified, performs targeted embodied experiments to test them, and uses the resulting discrepancies as data for model revision. Related LLM-assisted scientific workflows have begun to make hypothesis generation, experiment selection, tool use, and plan revision explicit in software agents \cite{boiko_autonomous_2023, lu_ai_2024, schmidgall_agent_2025, gottweis_towards_2025}; coupling those planning capabilities to robotic embodiment would extend this automation from in-silico reasoning to physical model probing.

In summary, our framework converts behavioral models into testable closed-loop predictions by training robotic policies in simulation and evaluating their transfer to live animal-robot interaction.
The resulting sim-to-real gaps provide a model-agnostic measure of model fidelity and a diagnostic tool for identifying behavioral features that require improved modeling. As biomimetic robots become increasingly capable, this approach offers a scalable pathway for validating and comparing interaction models, for uncovering model-specific behavioral signatures, and ultimately for advancing our mechanistic understanding of collective animal behavior.

\begin{figure}
	\centering
	\includegraphics[width=\textwidth]{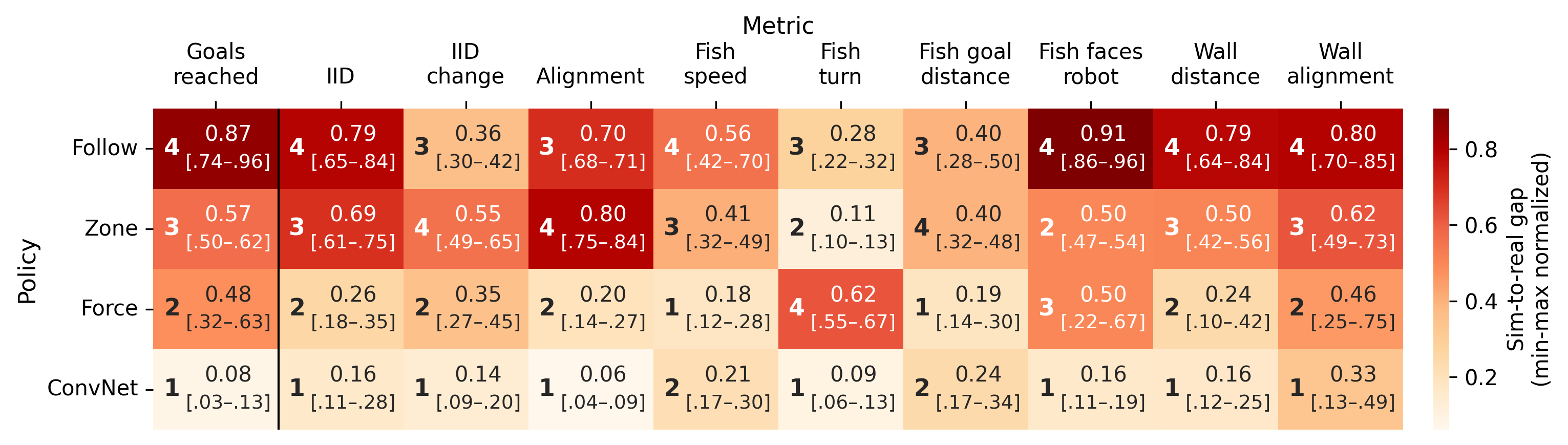}
	\caption{\textbf{Sim-to-real gap across behavioral metrics.}
	Annotated heatmap summarizing sim-to-real gaps for each policy (rows) and metric (columns). Each cell reports a point estimate of the sim-to-real gap based on trial-level Wasserstein distances, min–max normalized per metric, with the 95\% bootstrap confidence interval shown below in brackets. The bold numbers on the left of each cell indicate the within-metric ranking of policies based on this point estimate (1 = smallest gap).
	Averaging ranks across all metrics except \emph{goals reached} yields mean ranks of $\pi_{Follow}$: 3.56, $\pi_{Zone}$: 3.11, $\pi_{Force}$: 2.11, and $\pi_{ConvNet}$: 1.22, consistent with the ranking induced by the \emph{goals reached} metric alone.
	}
	\label{fig:gap}
\end{figure}

\begin{figure}
	\centering
	\includegraphics[width=0.6\textwidth]{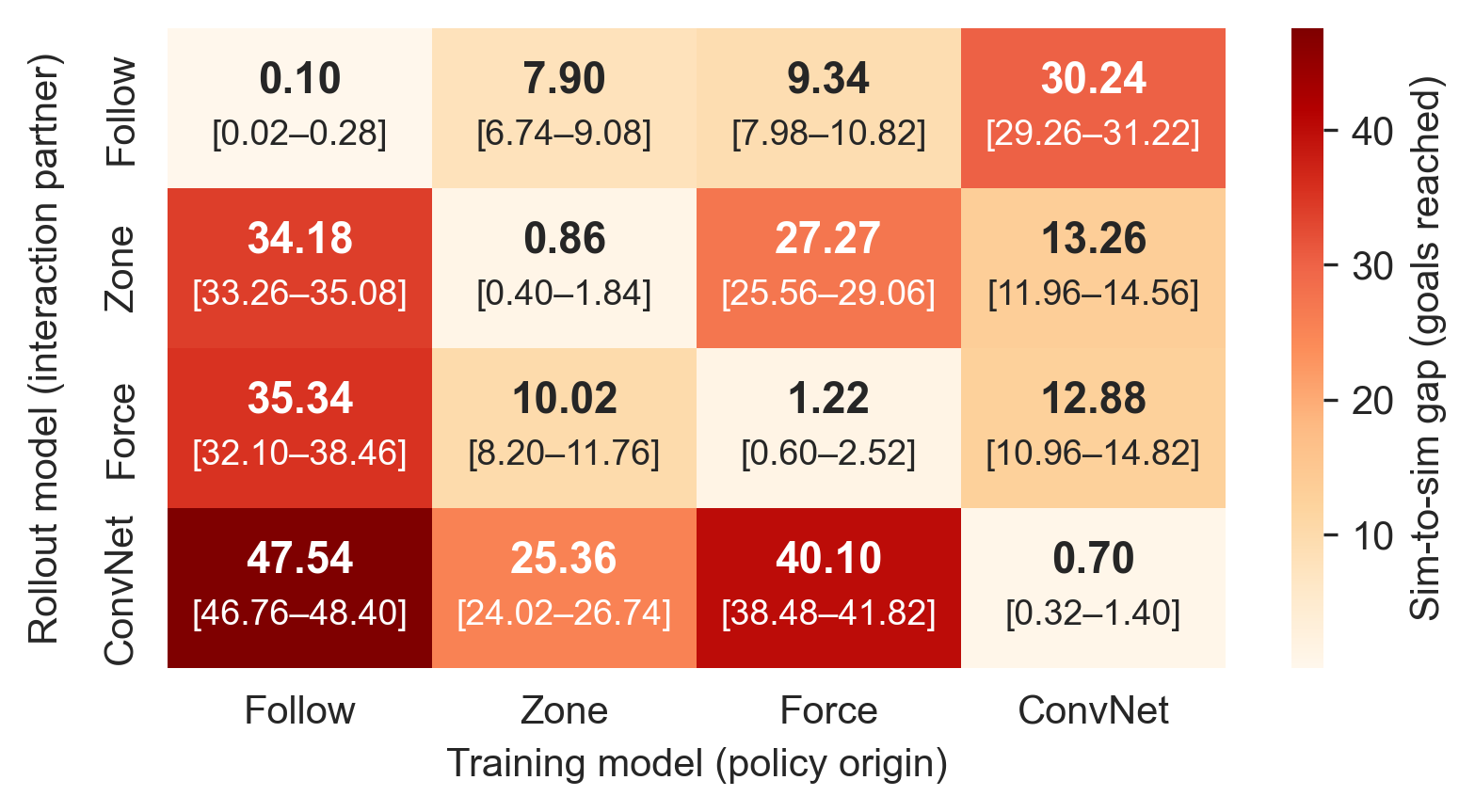}
	\caption{ \textbf{Sim-to-sim transfer between behavioral models.}
    Wasserstein distances (median and 95\% bootstrap CI) between the distributions of goals reached when a policy trained with a given behavioral model (“Training model”, columns) is rolled out using the same or a different model (“Rollout model”, rows). Each cell quantifies how well a policy trained on one model generalizes to another. Distances on the diagonal remain small but nonzero due to bootstrap variability, whereas off-diagonal distances are substantially larger, demonstrating that the four fish behavior models induce distinct interaction dynamics. Consequently, RL policies learn model-specific strategies that do not transfer across models, confirming that the models differ meaningfully in their behavioral assumptions.
    }
	\label{fig:simtosim}
\end{figure}

\begin{figure}
	\centering
	\includegraphics[width=0.8\textwidth]{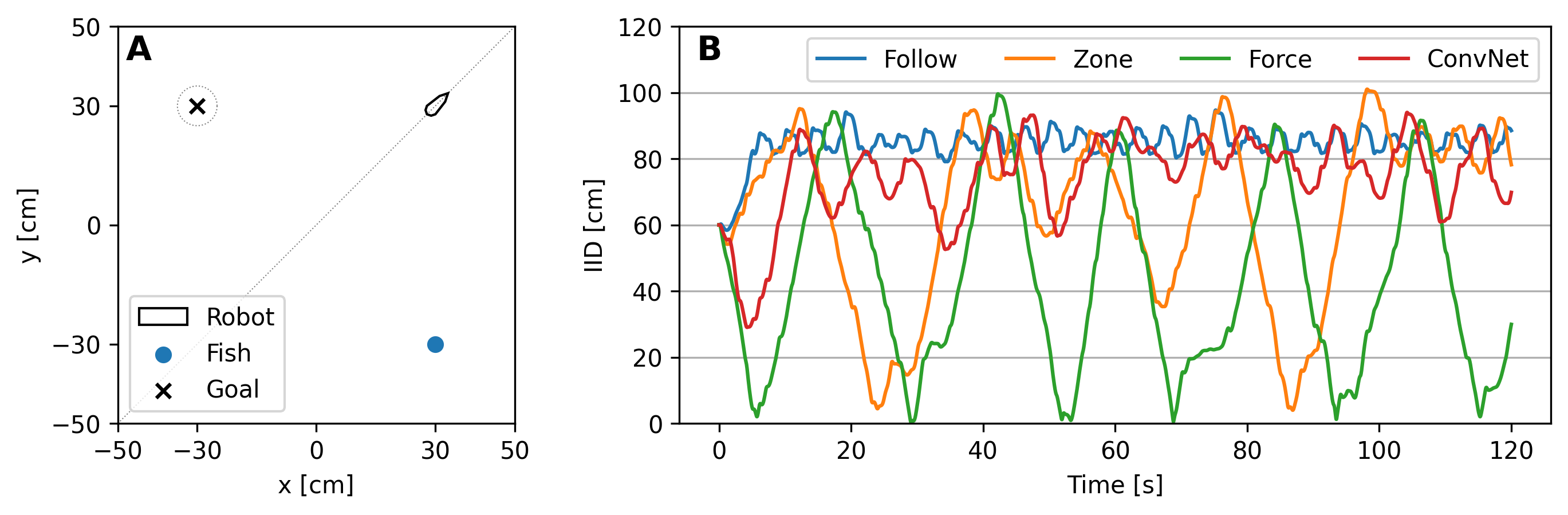}
	\caption{\textbf{Approach behavior toward a stationary fish model across policies.}
	(\textbf{A}) Top-down view of the initial setup, with the robot facing the center of the tank, the goal positioned on its right, and a stationary fish model placed on its left. 
	(\textbf{B}) Inter-individual distance (IID) over time. All policies except $\pi_{Follow}$ reduce their distance to the fish model to some extent. $\pi_{Force}$ repeatedly approaches very closely and spends more than half of the trial on the side of the tank nearest the fish model.}
	\label{fig:approach}
\end{figure}

\clearpage

\nocite{rl_data_zenodo,rf_rl_paper_code}
\bibliography{references}

@Misc{methods,
  note = {Materials and methods are available as supplementary material},
}

@misc{rl_data_zenodo,
	author = {Hocke, Mathis and Gerken, Andreas and Bierbach, David and Krause, Jens and Landgraf, Tim},
	title = {Data for the publication ``Robots that learn to evaluate models of collective behavior''},
	year = {2026},
	publisher = {Zenodo},
	doi = {10.5281/zenodo.19856850},
	url = {https://doi.org/10.5281/zenodo.19856850},
	note = {Data set},
}

@misc{rf_rl_paper_code,
	author = {Hocke, Mathis and Gerken, Andreas and Bierbach, David and Krause, Jens and Landgraf, Tim},
	title = {{rf\_rl\_paper}: Code for the publication ``Robots that learn to evaluate models of collective behavior''},
	year = {2026},
	url = {https://git.imp.fu-berlin.de/bioroboticslab/robofish/rl_paper},
	note = {Source code repository; accessed 29 April 2026},
}

@article{maxeiner_social_2023,
	title = {Social competence improves the performance of biomimetic robots leading live fish},
	volume = {18},
	issn = {1748-3190},
	url = {https://dx.doi.org/10.1088/1748-3190/acca59},
	doi = {10.1088/1748-3190/acca59},
	language = {en},
	number = {4},
	urldate = {2023-06-15},
	journal = {Bioinspiration \& Biomimetics},
	publisher = {IOP Publishing},
	author = {Maxeiner, Moritz and Hocke, Mathis and Moenck, Hauke J. and Gebhardt, Gregor H. W. and Weimar, Nils and Musiolek, Lea and Krause, Jens and Bierbach, David and Landgraf, Tim},
	month = may,
	year = {2023},
	pages = {045001},
}

@misc{klamser_impact_2021,
	title = {Impact of {Variable} {Speed} on {Collective} {Movement} of {Animal} {Groups}},
	url = {http://arxiv.org/abs/2106.00959},
	doi = {10.48550/arXiv.2106.00959},
	urldate = {2023-01-09},
	publisher = {arXiv},
	author = {Klamser, Pascal P. and Gómez-Nava, Luis and Landgraf, Tim and Jolles, Jolle W. and Bierbach, David and Romanczuk, Pawel},
	month = jun,
	year = {2021},
	note = {arXiv:2106.00959 [physics, q-bio]},
}

@article{faria_novel_2010,
	title = {A novel method for investigating the collective behaviour of fish: introducing ‘{Robofish}’},
	doi = {10.1007/s00265-010-0988-y},
	journal = {Behavioral Ecology and Sociobiology},
	author = {{Faria}},
	year = {2010},
}

@article{landgraf_robofish_2016,
	title = {{RoboFish}: increased acceptance of interactive robotic fish with realistic eyes and natural motion patterns by live {Trinidadian} guppies},
	volume = {11},
	issn = {1748-3190},
	shorttitle = {{RoboFish}},
	url = {https://doi.org/10.1088%2F1748-3190%2F11%2F1%2F015001},
	doi = {10.1088/1748-3190/11/1/015001},
	language = {en},
	number = {1},
	urldate = {2020-06-10},
	journal = {Bioinspiration \& Biomimetics},
	publisher = {IOP Publishing},
	author = {Landgraf, Tim and Bierbach, David and Nguyen, Hai and Muggelberg, Nadine and Romanczuk, Pawel and Krause, Jens},
	month = jan,
	year = {2016},
	note = {00066},
	pages = {015001},
}

@article{vicsek_novel_1995,
	title = {Novel {Type} of {Phase} {Transition} in a {System} of {Self}-{Driven} {Particles}},
	volume = {75},
	issn = {0031-9007, 1079-7114},
	url = {https://link.aps.org/doi/10.1103/PhysRevLett.75.1226},
	doi = {10.1103/PhysRevLett.75.1226},
	language = {en},
	number = {6},
	urldate = {2017-07-17},
	journal = {Physical Review Letters},
	author = {Vicsek, Tamás and Czirók, András and Ben-Jacob, Eshel and Cohen, Inon and Shochet, Ofer},
	month = aug,
	year = {1995},
	note = {04436},
	pages = {1226--1229},
}

@article{aoki_simulation_1982,
	title = {A simulation study on the schooling mechanism in fish.},
	volume = {48},
	issn = {1349-998X, 0021-5392},
	url = {http://joi.jlc.jst.go.jp/JST.Journalarchive/suisan1932/48.1081?from=CrossRef},
	doi = {10.2331/suisan.48.1081},
	language = {en},
	number = {8},
	urldate = {2017-07-17},
	journal = {NIPPON SUISAN GAKKAISHI},
	author = {Aoki, Ichiro},
	year = {1982},
	note = {00000},
	pages = {1081--1088},
}

@article{reynolds_flocks_1987,
	title = {Flocks, herds and schools: {A} distributed behavioral model},
	volume = {21},
	issn = {00978930},
	shorttitle = {Flocks, herds and schools},
	url = {http://portal.acm.org/citation.cfm?doid=37402.37406},
	doi = {10.1145/37402.37406},
	language = {en},
	number = {4},
	urldate = {2017-07-24},
	journal = {ACM SIGGRAPH Computer Graphics},
	author = {Reynolds, Craig W.},
	month = aug,
	year = {1987},
	note = {09308},
	pages = {25--34},
}

@article{couzinEffectiveLeadershipDecisionmaking2005,
	title = {Effective leadership and decision-making in animal groups on the move},
	volume = {433},
	issn = {0028-0836, 1476-4679},
	url = {http://www.nature.com/doifinder/10.1038/nature03236},
	doi = {10.1038/nature03236},
	number = {7025},
	urldate = {2017-06-22},
	journal = {Nature},
	author = {Couzin, Iain D. and Krause, Jens and Franks, Nigel R. and Levin, Simon A.},
	month = feb,
	year = {2005},
	note = {01743},
	pages = {513--516},
}

@article{polverino_fish_2013,
	title = {Fish and {Robots} {Swimming} {Together} in a {Water} {Tunnel}: {Robot} {Color} and {Tail}-{Beat} {Frequency} {Influence} {Fish} {Behavior}},
	doi = {10.1371/journal.pone.0077589},
	journal = {PLoS ONE},
	author = {{Polverino}},
	year = {2013},
}

@article{eyjolfsdottir_learning_2016,
	title = {Learning recurrent representations for hierarchical behavior modeling},
	url = {http://arxiv.org/abs/1611.00094},
	urldate = {2020-06-17},
	journal = {arXiv:1611.00094 [cs]},
	author = {Eyjolfsdottir, Eyrun and Branson, Kristin and Yue, Yisong and Perona, Pietro},
	month = nov,
	year = {2016},
	note = {arXiv: 1611.00094},
}

@article{krause_interactive_2011,
	title = {Interactive robots in experimental biology},
	volume = {26},
	issn = {01695347},
	url = {http://linkinghub.elsevier.com/retrieve/pii/S0169534711000851},
	doi = {10.1016/j.tree.2011.03.015},
	language = {en},
	number = {7},
	urldate = {2017-07-24},
	journal = {Trends in Ecology \& Evolution},
	author = {Krause, Jens and Winfield, Alan F.T. and Deneubourg, Jean-Louis},
	month = jul,
	year = {2011},
	note = {00112},
	pages = {369--375},
}

@article{heras_deep_2019,
	title = {Deep attention networks reveal the rules of collective motion in zebrafish},
	volume = {15},
	issn = {1553-7358},
	url = {https://dx.plos.org/10.1371/journal.pcbi.1007354},
	doi = {10.1371/journal.pcbi.1007354},
	language = {en},
	number = {9},
	urldate = {2020-07-08},
	journal = {PLOS Computational Biology},
	author = {Heras, Francisco J. H. and Romero-Ferrero, Francisco and Hinz, Robert C. and de Polavieja, Gonzalo G.},
	editor = {Battaglia, Francesco P.},
	month = sep,
	year = {2019},
	note = {00008},
	pages = {e1007354},
}

@article{tobin_domain_2017,
	title = {Domain {Randomization} for {Transferring} {Deep} {Neural} {Networks} from {Simulation} to the {Real} {World}},
	url = {http://arxiv.org/abs/1703.06907},
	urldate = {2020-06-24},
	journal = {arXiv:1703.06907 [cs]},
	author = {Tobin, Josh and Fong, Rachel and Ray, Alex and Schneider, Jonas and Zaremba, Wojciech and Abbeel, Pieter},
	month = mar,
	year = {2017},
	note = {00726 
arXiv: 1703.06907},
}

@article{blanco-mulero_benchmarking_2024,
	title = {Benchmarking the {Sim}-to-{Real} {Gap} in {Cloth} {Manipulation}},
	volume = {9},
	issn = {2377-3766},
	url = {https://doi.org/10.1109/LRA.2024.3360814},
	doi = {10.1109/LRA.2024.3360814},
	number = {3},
	journal = {IEEE Robotics and Automation Letters},
	publisher = {IEEE},
	author = {Blanco-Mulero, David and Barbany, Oriol and Alcan, Gokhan and Colome, Adria and Torras, Carme and Kyrki, Ville},
	year = {2024},
	pages = {2981--2988},
}

@article{brockman_openai_2016,
	title = {{OpenAI} {Gym}},
	url = {http://arxiv.org/abs/1606.01540},
	urldate = {2018-03-15},
	journal = {arXiv:1606.01540 [cs]},
	author = {Brockman, Greg and Cheung, Vicki and Pettersson, Ludwig and Schneider, Jonas and Schulman, John and Tang, Jie and Zaremba, Wojciech},
	month = jun,
	year = {2016},
	note = {01616 
arXiv: 1606.01540},
}

@misc{schulmanProximalPolicyOptimization2017,
        title = {Proximal {Policy} {Optimization} {Algorithms}},
        url = {http://arxiv.org/abs/1707.06347},
        doi = {10.48550/arXiv.1707.06347},
        urldate = {2026-03-19},
        publisher = {arXiv},
        author = {Schulman, John and Wolski, Filip and Dhariwal, Prafulla and Radford, Alec and Klimov, Oleg},
        month = aug,
        year = {2017},
        note = {arXiv:1707.06347 [cs]},
}

@inproceedings{landgraf_interactive_2013,
	address = {Berlin, Heidelberg},
	series = {Lecture {Notes} in {Computer} {Science}},
	title = {Interactive {Robotic} {Fish} for the {Analysis} of {Swarm} {Behavior}},
	isbn = {978-3-642-38703-6},
	doi = {10.1007/978-3-642-38703-6_1},
	language = {en},
	booktitle = {Advances in {Swarm} {Intelligence}},
	publisher = {Springer},
	author = {Landgraf, Tim and Nguyen, Hai and Forgo, Stefan and Schneider, Jan and Schröer, Joseph and Krüger, Christoph and Matzke, Henrik and Clément, Romain O. and Krause, Jens and Rojas, Raúl},
	editor = {Tan, Ying and Shi, Yuhui and Mo, Hongwei},
	year = {2013},
	note = {00019},
	pages = {1--10},
}

@article{couzin_collective_2002,
	title = {Collective {Memory} and {Spatial} {Sorting} in {Animal} {Groups}},
	volume = {218},
	issn = {0022-5193},
	url = {http://www.sciencedirect.com/science/article/pii/S0022519302930651},
	doi = {10.1006/jtbi.2002.3065},
	language = {en},
	number = {1},
	urldate = {2020-06-16},
	journal = {Journal of Theoretical Biology},
	author = {Couzin, IAIN D. and Krause, JENS and James, RICHARD and Ruxton, GRAEME D. and Franks, NIGEL R.},
	month = sep,
	year = {2002},
	note = {01717},
	pages = {1--11},
}

@article{bierbach_using_2018,
	title = {Using a robotic fish to investigate individual differences in social responsiveness in the guppy},
	volume = {5},
	url = {https://royalsocietypublishing.org/doi/full/10.1098/rsos.181026},
	doi = {10.1098/rsos.181026},
	number = {8},
	urldate = {2019-02-11},
	journal = {Royal Society Open Science},
	author = {Bierbach, David and Landgraf, Tim and Romanczuk, Pawel and Lukas, Juliane and Nguyen, Hai and Wolf, Max and Krause, Jens},
	year = {2018},
	pages = {181026},
}

@misc{bennettCurtaGeneralpurposeHighPerformance2020,
        title = {Curta: {A} {General}-purpose {High}-{Performance} {Computer} at {ZEDAT}, {Freie} {Universität} {Berlin}},
        shorttitle = {Curta},
        url = {https://refubium.fu-berlin.de/handle/fub188/26993},
        doi = {10.17169/REFUBIUM-26754},
        urldate = {2022-01-01},
        publisher = {Freie Universität Berlin},
        author = {Bennett, Loris and Melchers, Bernd and Proppe, Boris},
        collaborator = {Universitätsbibliothek Der FU Berlin and Universitätsbibliothek Der FU Berlin},
        year = {2020},
        note = {Artwork Size: 5 S.},
        pages = {5 S.},
}

@article{huth_simulation_1992,
	title = {The simulation of the movement of fish schools},
	volume = {156},
	issn = {0022-5193},
	url = {https://www.sciencedirect.com/science/article/pii/S0022519305806812},
	doi = {10.1016/S0022-5193(05)80681-2},
	number = {3},
	urldate = {2025-11-28},
	journal = {Journal of Theoretical Biology},
	author = {Huth, Andreas and Wissel, Christian},
	month = jun,
	year = {1992},
	pages = {365--385},
}

@inproceedings{peng_sim--real_2018,
	address = {Brisbane, QLD},
	title = {Sim-to-{Real} {Transfer} of {Robotic} {Control} with {Dynamics} {Randomization}},
	isbn = {978-1-5386-3081-5},
	url = {https://ieeexplore.ieee.org/document/8460528/},
	doi = {10.1109/ICRA.2018.8460528},
	urldate = {2025-11-28},
	booktitle = {2018 {IEEE} {International} {Conference} on {Robotics} and {Automation} ({ICRA})},
	publisher = {IEEE},
	author = {Peng, Xue Bin and Andrychowicz, Marcin and Zaremba, Wojciech and Abbeel, Pieter},
	month = may,
	year = {2018},
	pages = {3803--3810},
}

@article{costa_automated_2020,
	title = {Automated {Discovery} of {Local} {Rules} for {Desired} {Collective}-{Level} {Behavior} {Through} {Reinforcement} {Learning}},
	volume = {8},
	issn = {2296-424X},
	url = {https://www.frontiersin.org/journals/physics/articles/10.3389/fphy.2020.00200/full},
	doi = {10.3389/fphy.2020.00200},
	language = {English},
	urldate = {2025-11-28},
	journal = {Frontiers in Physics},
	publisher = {Frontiers},
	author = {Costa, Tiago and Laan, Andres and Heras, Francisco J. H. and de Polavieja, Gonzalo G.},
	month = jun,
	year = {2020},
}

@article{papaspyrosQuantifyingbiomimicrygap2024,
        title = {Quantifying the biomimicry gap in biohybrid robot-fish pairs},
        volume = {19},
        issn = {1748-3190},
        url = {https://doi.org/10.1088/1748-3190/ad577a},
        doi = {10.1088/1748-3190/ad577a},
        language = {en},
        number = {4},
        urldate = {2026-03-03},
        journal = {Bioinspiration \& Biomimetics},
        publisher = {IOP Publishing},
        author = {Papaspyros, Vaios and Theraulaz, Guy and Sire, Clément and Mondada, Francesco},
        month = jun,
        year = {2024},
        pages = {046020},
}

@article{halloySocialIntegrationRobots2007,
	title = {Social {Integration} of {Robots} into {Groups} of {Cockroaches} to {Control} {Self}-{Organized} {Choices}},
	volume = {318},
	issn = {0036-8075, 1095-9203},
	url = {http://www.sciencemag.org/cgi/doi/10.1126/science.1144259},
	doi = {10.1126/science.1144259},
	language = {en},
	number = {5853},
	urldate = {2017-07-24},
	journal = {Science},
	author = {Halloy, J. and Sempo, G. and Caprari, G. and Rivault, C. and Asadpour, M. and Tache, F. and Said, I. and Durier, V. and Canonge, S. and Ame, J. M. and Detrain, C. and Correll, N. and Martinoli, A. and Mondada, F. and Siegwart, R. and Deneubourg, J. L.},
	month = nov,
	year = {2007},
	note = {00336},
	pages = {1155--1158},
}

@misc{landgrafDancingHoneyBee2018c,
        title = {Dancing {Honey} bee {Robot} {Elicits} {Dance}-{Following} and {Recruits} {Foragers}},
        url = {http://arxiv.org/abs/1803.07126},
        doi = {10.48550/arXiv.1803.07126},
        urldate = {2026-04-27},
        publisher = {arXiv},
        author = {Landgraf, Tim and Bierbach, David and Kirbach, Andreas and Cusing, Rachel and Oertel, Michael and Lehmann, Konstantin and Greggers, Uwe and Menzel, Randolf and Rojas, Raúl},
        month = mar,
        year = {2018},
        note = {arXiv:1803.07126 [cs]},
}

@article{boiko_autonomous_2023,
        title = {Autonomous chemical research with large language models},
        volume = {624},
        url = {https://www.nature.com/articles/s41586-023-06792-0},
        doi = {10.1038/s41586-023-06792-0},
        journal = {Nature},
        author = {Boiko, Daniil A. and MacKnight, Robert and Kline, Ben and Gomes, Gabe},
        year = {2023},
        pages = {570--578},
}

@misc{lu_ai_2024,
        title = {The {AI} {Scientist}: {Towards} {Fully} {Automated} {Open}-{Ended} {Scientific} {Discovery}},
        url = {https://arxiv.org/abs/2408.06292},
        doi = {10.48550/arXiv.2408.06292},
        author = {Lu, Chris and Lu, Cong and Lange, Robert Tjarko and Foerster, Jakob and Clune, Jeff and Ha, David},
        year = {2024},
        note = {arXiv:2408.06292 [cs]},
}

@misc{schmidgall_agent_2025,
        title = {Agent {Laboratory}: {Using} {LLM} {Agents} as {Research} {Assistants}},
        url = {https://arxiv.org/abs/2501.04227},
        doi = {10.48550/arXiv.2501.04227},
        author = {Schmidgall, Samuel and Su, Yusheng and Wang, Ze and Sun, Ximeng and Wu, Jialian and Yu, Xiaodong and Liu, Jiang and Moor, Michael and Liu, Zicheng and Barsoum, Emad},
        year = {2025},
        note = {arXiv:2501.04227 [cs]},
}

@misc{gottweis_towards_2025,
        title = {Towards an {AI} co-scientist},
        url = {https://arxiv.org/abs/2502.18864},
        doi = {10.48550/arXiv.2502.18864},
        author = {Gottweis, Juraj and Weng, Wei-Hung and Daryin, Alexander and Tu, Tao and Palepu, Anil and Sirkovic, Petar and Myaskovsky, Artiom and Weissenberger, Felix and Rong, Keran and Tanno, Ryutaro and Saab, Khaled and others},
        year = {2025},
        note = {arXiv:2502.18864 [cs]},
}

@article{krauseLeadershipfishshoals2000,
	title = {Leadership in fish shoals},
	volume = {1},
	issn = {1467-2979},
	url = {https://onlinelibrary.wiley.com/doi/10.1111/j.1467-2979.2000.tb00001.x},
	doi = {10.1111/j.1467-2979.2000.tb00001.x},
	language = {en},
	number = {1},
	urldate = {2026-04-27},
	journal = {Fish and Fisheries},
	publisher = {John Wiley \& Sons, Ltd},
	author = {Krause, J. and Hoare, D. and Krause, S. and Hemelrijk, C. K. and Rubenstein, D. I.},
	month = mar,
	year = {2000},
	pages = {82--89},
}

@article{strandburg-peshkinInferringInfluenceLeadership2018a,
	title = {Inferring influence and leadership in moving animal groups},
	volume = {373},
	url = {https://royalsocietypublishing.org/doi/10.1098/rstb.2017.0006},
	doi = {10.1098/rstb.2017.0006},
	number = {1746},
	urldate = {2020-08-12},
	journal = {Philosophical Transactions of the Royal Society B: Biological Sciences},
	publisher = {Royal Society},
	author = {Strandburg-Peshkin, Ariana and Papageorgiou, Danai and Crofoot, Margaret C. and Farine, Damien R.},
	month = may,
	year = {2018},
	note = {00032},
	pages = {20170006},
}
\bibliographystyle{sciencemag}

\section*{Acknowledgments}
We thank Gregor Gebhardt and Julian Stastny for contributions during the early stages of the project, and Janosch Brandhorst for helpful discussions, proofreading, and feedback on the manuscript.
The authors would like to thank the HPC Service of FUB-IT, Freie Universität Berlin, for computing time \cite{bennettCurtaGeneralpurposeHighPerformance2020}.

\paragraph*{Funding:}
We acknowledge financial support from the German Research Foundation (BI 1828/2-1, LA 3534/1-1) and Germany's Excellence Strategy (EXC 2002/1 'Science of Intelligence', Project Number 390523135). Furthermore, Mathis Hocke and Andreas Gerken were supported by the Elsa-Neumann Scholarship of the State of Berlin.

\paragraph*{Competing interests:}
There are no competing interests to declare.

\paragraph*{Data and materials availability:}

All data and software required to reproduce the results are archived in a public Zenodo record and made available without embargo at \url{https://doi.org/10.5281/zenodo.19856850} \cite{rl_data_zenodo}. The archive is listed as Data S1 and contains rollout data, trained policy checkpoints, policy-training logs, full policy hyperparameter files, and approach-assay data. Rendered videos from the archived rollouts are provided for convenience in a YouTube playlist at \url{https://youtube.com/playlist?list=PLs7Vp-pCDX7xKo9AqVq4Bfw195wRdbmMH}. The code is available in the repository at \url{https://git.imp.fu-berlin.de/bioroboticslab/robofish/rl_paper} \cite{rf_rl_paper_code}.

\subsection*{Supplementary materials}
Materials and Methods\\
Tables S1 to S3\\
References \textit{(34)}\\
Data S1

\newpage

\renewcommand{\thefigure}{S\arabic{figure}}
\renewcommand{\thetable}{S\arabic{table}}
\renewcommand{\theequation}{S\arabic{equation}}
\renewcommand{\thepage}{S\arabic{page}}
\setcounter{figure}{0}
\setcounter{table}{0}
\setcounter{equation}{0}
\setcounter{page}{1}

\begin{center}
\section*{Supplementary Materials for\\ \scititle}

	Mathis~Hocke$^{\ast}$,
	Andreas~Gerken,
    David~Bierbach,
    Jens~Krause,
	Tim~Landgraf$^{\ast}$\\
	\small$^\ast$Corresponding authors. Email: mathis.hocke@fu-berlin.de, tim.landgraf@fu-berlin.de
\end{center}

\subsubsection*{This PDF file includes:}
Materials and Methods\\
Tables S1 to S3\\ 
References (35)\\
\subsubsection*{Other Supplementary Materials for this manuscript:}
Data S1

\newpage

\subsection*{Materials and Methods}

\subsubsection*{RoboFish Platform and Experimental Arena}
The RoboFish setup followed the design of \cite{maxeiner_social_2023}. The experimental arena consisted of a 100~cm $\times$~100~cm tank filled with 7~cm of aged tap water. A biomimetic fish replica was magnetically actuated from below using a two-wheeled robotic platform. A top-mounted camera (resolution: 2048 $\times$ 2048, frame rate: 25~Hz) recorded trajectories of both fish and robot.

At the start of each trial, the robot performed a standardized milling motion for two minutes at the center of the tank to habituate the fish. The test fish was initially held in a cylindrical container in a corner of the tank. The container was removed after one minute, and after another minute the robot switched from milling to closed-loop policy control. The robot received one turn-angle command per second from the RL policy. Low-level motion (deceleration, turning, acceleration) was governed by a PID controller. Maximum translational speed was 15~cm/s and maximum turning speed was 180°/s. Trials lasted 15~min, during which the robot attempted to lead the fish alternately between two goal locations positioned symmetrically in opposite corners, each 30~cm from the walls.

\subsubsection*{Reinforcement Learning Framework}
Policies were trained in simulation using the GuppyGym environment, built on the OpenAI Gym API \cite{brockman_openai_2016}. The virtual environment matched the physical tank dimensions and included a simulated robot, fish model, and circular goal region (radius 5~cm).
Policies were trained using Proximal Policy Optimization (PPO) \cite{schulmanProximalPolicyOptimization2017} with the following hyperparameters:
learning rate 0.00026,
batch size 8192,
minibatch size 512,
PPO epochs 100,
discount factor $\gamma$=0.9453,
GAE parameter $\lambda$=0.9652,
entropy coefficient 0,
value-function coefficient 1,
and clipping parameter $\epsilon$=0.1112.
Training ran for 832300 environment steps using Ray RLlib v2.0.1.
Full training configurations and hyperparameters are archived in Data S1 in the policy-specific \texttt{params.json} files, together with the trained checkpoints, \texttt{progress.csv} training logs and \texttt{result.json} summaries.

Episodes lasted 41 s during training and 15 min during evaluation. During training, the goal position was randomized at episode start and each time the robot reached it. During evaluation, the goal switched only when reached, alternating deterministically between two fixed corners of the arena.

The robot's observation space combined spatial information about the environment, the fish, and the goal.
It included 36 wall raycasts providing normalized distances to the tank boundaries,
36 angular fish bins encoding the normalized distance to the fish (set to zero when the fish was outside the corresponding bin),
and 36 angular goal bins encoding the normalized distance to the goal location.
The wall raycasts and angular bins were distributed evenly across a full 360° field of view.
In addition, the observation vector contained the previous fish and goal observations.

The action space was one-dimensional and continuous, representing a relative turn angle (radians), clipped to [-$\pi$,$\pi$].

Rewards followed:
$$r_t = r_{goal} + p_{wall},$$
where $r_{goal}=1$ if the fish entered the goal area (otherwise 0).
A wall-avoidance penalty $p_{wall}=-0.1$ was applied whenever the robot was within 3 cm of the tank boundary.

\subsubsection*{Behavioral Models}
All behavioral models are agent-based and determine future movement (linear velocity $v$, angular velocity $\omega$), based on the current state of the agents. The models can be used autoregressively to generate behavior over multiple timesteps.

\begin{description}
    \item[$M_{Follow}$] As a baseline we use an agent model that moves at constant speed and always rotates so that it points directly to the closest other agent. This leads to simple aggregation and the ability to follow another agent.

    \item[$M_{Zone}$] The model described in \cite{couzin_collective_2002} has constant speed and three mathematically described behaviors which are selected depending on the distance between agents using thresholds. In the closest zone, the agents repulse each other (zone of repulsion), in the next zone they align their orientations (zone of orientation), and in the largest zone they attract each other (zone of attraction). The model was extended by controllable noise added to $v$ and $\omega$, as well as by a wall detection module. If the agent is close to a wall and is heading towards the wall, a repulsion force is added. If the agent is in a certain distance of the wall and is heading away from the wall, an attraction force is added. This leads to the ability to swim close to walls while avoiding collisions, which is a behavior we found often in real swarm recordings.

    \item[$M_{Force}$] The variable-speed model described in \cite{klamser_impact_2021} has a preferred speed that it tries to reach. Additionally it has a preferred distance to other agents, which leads to attraction if agents are far apart and repulsion if they are close. These forces can influence the velocity of the agents. The model was also modified to introduce nondeterministic behavior and wall repulsion.

    \item[$M_{ConvNet}$] The convolutional model is a neural agent trained via behavioral cloning on live fish recordings as training data. The agent gets an egocentric view of the environment as input and has the task to predict the corresponding action. The input is built from two components: wall distances sampled by raycasts at uniform angular increments over a full 360° field of view, and nearest-neighbor information encoded in equal angular sectors, each storing the distance and relative angle of the closest agent. The agent consists of a neural network with multiple layers to predict the outputs. Since adjacent raycast and sector values have a spatial relationship, the first layer of the network is a convolutional layer, followed by two fully connected layers with Leaky ReLU activation functions. The action distribution is a 2D grid discretization of $v$ and $\omega$. The action is selected by a final softmax layer and probabilistic sampling afterwards. The model is trained using behavioral cloning by minimizing the negative log-likelihood of given live fish state action pairs.
\end{description}

\subsubsection*{Sim-to-Real Gap Computation}
For each policy, we collected distributions of goals reached in simulation (18 episodes) and in 18 physical trials. The sim-to-real gap was quantified using the 1-Wasserstein distance between these empirical distributions.
We repeated this procedure for additional behavioral metrics to test the generality of the ranking:

\begin{description}
    \item[Goals reached:] Number of goals reached in 15~min.
    \item[IID:] Inter-individual distance (between robot and fish).
    \item[IID change:] $\Delta\text{IID}/\Delta t$.
    \item[Alignment:] Dot product of robot and fish orientations.
    \item[Fish speed:] Linear velocity of the fish.
    \item[Fish turn:] Angular velocity of the fish.
    \item[Fish goal distance:] Euclidean distance from fish to goal.
    \item[Fish faces robot:] The dot product between the fish’s unit orientation vector and the normalized vector pointing from the fish to the robot (1 = directly facing, 0 = perpendicular, -1 = facing away).   
    \item[Wall distance:] Distance from fish to nearest wall.
    \item[Wall alignment:] Dot product of fish orientation and nearest wall orientation.
\end{description}
All metrics were computed at the tracking frame rate (25~Hz), except for goals reached, which was computed on a per-trial basis.

\subsubsection*{Sim-to-Sim Transfer}
To evaluate model–policy specificity, each trained policy was tested against each fish model in simulation for 50 episodes, producing an $N\times N$ matrix of distributions of the number of goals reached, where N = 4. Large performance drops indicated model–policy mismatch, reflecting both differences in learned policies and underlying behavioral models. 

\subsubsection*{Statistical Analysis}

    Analysis code and a fully specified Python environment are available at \url{https://git.imp.fu-berlin.de/bioroboticslab/robofish/rl_paper} \cite{rf_rl_paper_code}. We used nonparametric, distribution-based methods throughout, motivated by small sample sizes and non-Gaussian behavioral distributions.

\paragraph*{Wasserstein distance and confidence intervals.}
To quantify differences between empirical behavioral distributions, we used the 1-Wasserstein distance.
For goals reached, we report the observed distance $W_{\mathrm{obs}}$ with a 95\% percentile-bootstrap confidence interval ($B=5000$) obtained by resampling trials with replacement within each environment and recomputing $W$.
For sim-to-sim transfer (goals reached), we report the median of the bootstrap distribution with its 95\% percentile-bootstrap confidence interval.

\paragraph*{Permutation tests.}
To test whether two distributions differed significantly, we used one-sided (upper-tail) permutation tests on
the Wasserstein distance:

\[
p = \Pr\!\left( W_{\mathrm{perm}} \ge W_{\mathrm{obs}} \right),
\]

where $W_{\mathrm{perm}}$ denotes the Wasserstein distance computed after randomly permuting group labels across the pooled samples (preserving group sizes) with the permutation repeated $10^5$ times and $W_{\mathrm{obs}}$ is the observed distance between the original groups.
Permutation tests were used for:
(i) significance tests for sim-to-real gaps ($N_{goals}$),
(ii) planned comparisons between each policy’s model and all other models in sim-to-sim transfer ($N_{goals}$).
All $p$-values in planned comparisons were Holm-corrected.

\paragraph*{Effect sizes.}
For two-sample, trial-level distributions, we report Cliff’s delta:

\[
\delta = \Pr(x > y) - \Pr(x < y),
\]

which is robust to unequal variances and non-Gaussian distributions.  
Magnitudes follow common thresholds: negligible $(|\delta| < 0.147)$, small $(<0.33)$, medium $(<0.474)$, and large otherwise.  
Effect sizes were computed for sim-to-real gaps (goals reached).

\paragraph*{Sim-to-sim model specificity tests.}
We constructed an $N \times N$ matrix ($N=4$ models) of bootstrap Wasserstein distances comparing the policy–model pairing (train model = rollout model) against all cross-model rollouts.  
For each policy, we performed three planned comparisons between the $N_{goals}$ distributions in the diagonal cell in Fig.~\ref{fig:simtosim} (train model = rollout model) and the three off-diagonal alternatives (train model $\neq$ rollout model).
All comparisons were significant after Holm correction (all $p<0.001$), indicating that policies learned strongly model-specific strategies.

We additionally performed Kruskal–Wallis omnibus tests for each trained policy on $N_{goals}$ distributions across rollout models.

\paragraph*{Sim-to-real gap (per-trial metrics).}
For the goals-reached metric, each policy had 18 simulated and 18 real trials.  
We computed:
\begin{enumerate}
\item the bootstrap CI for the Wasserstein gap,
\item a permutation test (100,000 permutations), and
\item Cliff’s delta.
\end{enumerate}
All policies showed highly significant sim-to-real differences (all $p<0.01$).

\paragraph*{Sim-to-real gap (per-time-step metrics).}
Per-time-step values (25\,Hz) are autocorrelated; therefore, Wasserstein distances were computed at the \textit{trial level} rather than by pooling all time steps.  
For each policy and metric, we compared each simulated trial with each real trial, yielding an $18 \times 18$ matrix $D$ of pairwise distances.
Because each trial appears in multiple pairwise comparisons, entries of $D$ are not independent; we therefore summarize the gap by the median of all matrix entries and estimate a 95\% confidence interval using a two-way cluster bootstrap that resamples simulated trials (rows) and real trials (columns) with replacement.
For Fig.~\ref{fig:gap}, values were min–max normalized within each metric.

\subsubsection*{Approach Behavior Assay}
Policies were evaluated in a simulated environment containing a stationary fish model. The robot began in a corner facing the tank center; the fish model and goal were placed at equal distances (30~cm from walls) to the left and right. Over 2-minute episodes, we measured minimum IID, mean IID, and the fraction of time the robot was closer to the fish model than the goal. See Fig.~\ref{fig:approach}.

\paragraph*{Temporal-trend analysis.}
To test whether performance changed over the 15-min control phase, we computed goals reached per minute (15 one-minute bins per trial) and fit a negative-binomial generalized linear model with a log link and time (minute index) as a predictor. Standard errors were clustered by trial to account for within-trial dependence across minute bins. We report the per-minute rate ratio and its 95\% CI (Table~\ref{tab:temporal}). In real experiments, we observed a small but statistically significant decline; in simulation, we observed a slight increase consistent with early-trial transients.

\paragraph*{Policy-retraining stability.}
To ensure that the observed sim-to-real differences reflect model-induced interaction dynamics rather than optimization noise, we retrained each policy five times with identical hyperparameters but independent random seeds.
Each trained policy was evaluated for 50 episodes on its training model and we recorded goals reached per episode.
We then decomposed variability in mean goals reached into (i) within-model (policy-to-policy) variance across the 6 policies per model (original + 5 retrains) and (ii) between-model variance across the four behavioral models.
We report the variance ratio $R = \sigma^2_{\mathrm{between}} / \sigma^2_{\mathrm{within}}$, where $\sigma^2_{\mathrm{within}}$ is the average (across models) of the variance in per-policy means within each model.
Uncertainty in $R$ was estimated by bootstrap resampling episodes within each policy (2{,}000 replicates), and a permutation test (5{,}000 permutations) shuffled per-policy means across model labels while preserving the number of policies per model (6).

\begin{table}
	\centering
	\caption{\textbf{Parameters for $M_{Zone}$.}
		}
	\label{tab:couzin}

    \vspace{1em}
    \begin{tabular}{l r}
	\hline
    \textbf{Parameter} & \textbf{value} \\
	\hline
    zone of repulsion & 2 cm \\
    zone of orientation & 6 cm \\
    zone of attraction & 10 cm \\
    speed & 6.30 cm/s \\
    field of view & $2\pi$ \\
    turn noise & 0.2 \\
    max angular velocity & 10 rad/s \\
	smoothing factor & 1 \\
    wall repulsion radius & 8.39 cm \\
    wall repulsion reaction angle & 1.88 \\
    wall repulsion strength & 40.51 \\
    \hline
    \end{tabular}
\end{table}

\begin{table}
	\centering
	\caption{\textbf{Parameters of $M_{Force}$.}}
	\label{tab:klamser}

	\vspace{1em}

	\begin{tabular}{l r}
	\hline
	\textbf{Parameter} & \textbf{Value} \\
	\hline
	\multicolumn{2}{l}{\textit{Social interaction parameters}} \\
	Interaction range (neighbor radius) & 15 cm \\
	Preferred inter-individual distance ($r_d$) & 5.0 cm \\
	Velocity alignment strength ($\mu_{\mathrm{alg}}$) & 0.1 \\
	Distance regulation strength ($\mu_d$) & 0.3 \\
	Distance interaction slope ($m_d$) & 0.2 \\
	\\
	\multicolumn{2}{l}{\textit{Individual motion parameters}} \\
	Preferred speed ($v_0$) & 0.2 cm / time step\\
	Speed relaxation rate ($\beta$) & 0.5 \\
	Rotational friction coefficient ($\alpha$) & 1.0 \\
	Angular noise intensity ($D_\varphi$) & 0.3 \\
	Speed noise intensity ($D_v$) & 0.2 \\
	\\
	\multicolumn{2}{l}{\textit{Wall interaction parameters (model extension)}} \\
	Wall angle factor (outward component) & -0.7 \\
	Wall angle factor (perpendicular component) & 0.3 \\
	Wall attraction strength & 0.3 \\
	Wall repulsion range & 11.0 \\
	Wall repulsion strength & 0.7 \\
	\hline
	\end{tabular}
\end{table}

\begin{table}
	\centering
    \caption{\textbf{Temporal trend in goals reached per minute.}
    We fit a negative-binomial regression (log link) of per-minute goal counts on time (minute index), separately for each policy and pooled across policies, using trial-clustered robust standard errors. The table reports the per-minute rate ratio (RR) with 95\% CI and the $p$-value for the time coefficient.
    }
	\label{tab:temporal}

    \vspace{1em}
    \begin{tabular}{l l c c}
	\hline
    \textbf{Env} & \textbf{Policy} & \textbf{RR/min (95\% CI)} & \textbf{p} \\
    \hline
    sim & total   & 1.006 [1.002, 1.010] & 0.004 \\
    sim & $\pi_{Follow}$  & 1.002 [1.000, 1.004] & 0.011 \\
    sim & $\pi_{Zone}$  & 1.008 [0.998, 1.018] & 0.107 \\
    sim & $\pi_{Force}$ & 1.009 [1.000, 1.018] & 0.056 \\
    sim & $\pi_{ConvNet}$ & 1.006 [0.984, 1.030] & 0.588 \\
    \hline
    real & total   & 0.974 [0.956, 0.992] & 0.005 \\
    real & $\pi_{Follow}$  & 0.965 [0.925, 1.006] & 0.092 \\
    real & $\pi_{Zone}$  & 0.943 [0.898, 0.989] & 0.016 \\
    real & $\pi_{Force}$ & 0.995 [0.964, 1.027] & 0.754 \\
    real & $\pi_{ConvNet}$ & 0.960 [0.939, 0.982] & 0.001 \\
    \hline
    \end{tabular}
\end{table}

\clearpage

\paragraph{Caption for Data S1.}
\textbf{Archived rollout data, trained policies, and approach-assay data.}
Data S1 is the Zenodo archive associated with this manuscript \cite{rl_data_zenodo}. It contains three top-level directories: \texttt{rollouts}, \texttt{policies}, and \texttt{approach\_test}. The \texttt{rollouts} directory contains one subdirectory per policy--rollout condition, with directory names of the form \texttt{Policy\_\_RolloutModel}; \texttt{Real} denotes live-fish experiments, and all other rollout-model labels denote simulated fish models. Each rollout subdirectory contains the corresponding HDF5 rollout files. The \texttt{policies} directory contains one subdirectory per trained policy (\texttt{Follow}, \texttt{Zone}, \texttt{Force}, and \texttt{ConvNet}), including the policy checkpoints, full hyperparameter and configuration files (\texttt{params.json}), training logs (\texttt{progress.csv}), and training-result summaries (\texttt{result.json}). The \texttt{approach\_test} directory contains HDF5 files for the stationary-fish approach assay for each policy.

\end{document}